\documentclass{article}

\usepackage[dblblindworkshop, final]{neurips_2026}
\usepackage[utf8]{inputenc} % allow utf-8 input
\usepackage[T1]{fontenc}    % use 8-bit T1 fonts
\usepackage{hyperref}       % hyperlinks
\usepackage{url}            % simple URL typesetting
\usepackage{booktabs}       % professional-quality tables
\usepackage{amsfonts}       % blackboard math symbols
\usepackage{nicefrac}       % compact symbols for 1/2, etc.
\usepackage{microtype}      % microtypography
\usepackage{xcolor}         % colors
\usepackage{amsmath}
\usepackage{amssymb}   
\usepackage{graphicx}
\usepackage{booktabs}
\usepackage{multirow}
\usepackage{placeins}

\newcommand{\method}{UniCR}
\title{Interventional Analysis of Evidence Faithfulness in Generative Cloud Removal}

\workshoptitle{XAI4Science: Knowledge Discovery and Trust through Interpretable Foundation Models}

\author{%
  \textbf{Yuxi Lu}$^{1}$ \quad
  \textbf{Kunqi Li}$^{1}$ \quad
  \textbf{Zhidong Li}$^{1}$ \quad
  \textbf{Chenya Huang}$^{1}$ \\
  \textbf{Biao Wu}$^{1}$ \quad
  \textbf{Qizhen Lan}$^{2}$ \quad
  \textbf{Ronghao Chen}$^{3}$ \quad
  \textbf{Bin Liang}$^{1}$\thanks{Corresponding author: \texttt{Bin.Liang@uts.edu.au}.} \\
  $^{1}$University of Technology Sydney \\
  $^{2}$University of Texas Health Science Center at Houston \\
  $^{3}$QuantaAlpha
}

\begin{document}

\maketitle

\begin{abstract} Cloud-free reconstruction from heavily obscured satellite imagery is inherently ambiguous. Task-specific restoration models may produce conservative or incomplete outputs, whereas a strong pretrained generative prior enables sharper and more coherent completion but can also introduce visually plausible structures that are inconsistent with the observed scene. We present \method{}, a generative cloud-removal framework that constrains a pretrained BAGEL backbone using explicit, source-specific optical, spectral, SAR, and temporal condition-token blocks. After establishing strong performance across multichannel, multimodal, and multitemporal benchmarks, we analyze how the model uses this heterogeneous evidence. A structural audit shows that strong-prior restoration changes the failure profile: it reduces missing structures relative to specialist baselines, but may introduce more prediction-only structures. Scene-correspondence interventions further show that \method{} functionally depends on matched NIR, spectral, SAR, and temporal observations rather than merely the presence of additional tokens. At the spatial level, intervention-derived patch attribution identifies spatial evidence groups whose replacement causes substantially greater degradation than random replacement, validating the faithfulness of the attribution ranking. Increasing heterogeneous evidence also consistently reduces prediction-only and missing structures. Nevertheless, residual reference-inconsistent content remains possible under severe ambiguity. These results show that explicit heterogeneous evidence can constrain, but cannot fully eliminate, unreliable completion by a strong generative prior. \end{abstract}

\section{Introduction}
\label{sec:intro}

Optical satellite imagery supports land-cover analysis, environmental monitoring, agricultural assessment, and disaster response~\cite{yuan2020deep,lulc_review_2023_isprs,ecosystem_dynamics_2022,agri_meta_review_2020,joyce_disaster_review_2009}. However, cloud contamination obscures land-surface structures and distorts spectral measurements, reducing the availability and reliability of optical observations~\cite{cloudremovalsurvey2025jstars}. In heavily obscured regions, cloud-free reconstruction is inherently underdetermined: multiple outputs may appear visually plausible, while only some remain consistent with the underlying scene.

Deep learning has substantially advanced cloud removal through conditional GANs, diffusion models, and task-specific restoration architectures~\cite{filmycloud2017cvprw,meraner2020cloud,zou2024diffcr,liu2025effective}. Strong pretrained generative models offer an additional advantage by providing rich structural and semantic priors for reconstructing severely obscured content. However, this capability also changes the reliability problem. Our structural audit shows that, compared with specialist restoration baselines, the strong generative prior reduces missing structures but produces more prediction-only structures. In other words, stronger completion can shift the failure profile from conservative omission toward a greater risk of visually plausible but reference-inconsistent completion.

Remote sensing provides complementary observations that can constrain this prior, including auxiliary optical bands, SAR measurements, and multitemporal acquisitions~\cite{filmycloud2017cvprw,glfcr2022isprs,ebel2023uncrtaints}. Existing methods commonly integrate such information through concatenation or specialized fusion modules coupled to a particular generator~\cite{meraner2020cloud,glfcr2022isprs,liu2025effective}. While effective for restoration, these entangled designs make it difficult to determine which evidence sources actually influence generation and whether apparently important evidence is functionally used. We therefore study heterogeneous cloud removal as a testbed for understanding how a strong pretrained generative model uses scientific observational evidence.

To this end, we propose \textbf{\method}, which combines a pretrained BAGEL backbone~\cite{deng2025emerging} with explicit source-specific condition-token blocks. As illustrated in Fig.~\ref{fig:unicr_framework}, modality-specific encoders and connectors project RGB, non-RGB optical, SAR, and temporal observations into packed condition tokens that interact with latent generation states throughout flow prediction. This representation retains source identity and spatial organization, allowing the same evidence interface to support both generation and controlled intervention. Rather than treating raw attention or gradients as explanations, we ask three questions: \emph{what} observation sources functionally constrain generation, which spatial evidence groups exert functional influence within a source, and \emph{whether} the identified dependence is faithful under intervention.

For source-level analysis, RGB serves as the anchor optical observation, while auxiliary NIR, spectral, SAR, and temporal evidence is tested using scene-correspondence interventions. We replace matched evidence with observations from other scenes, preserving modality type and token structure while breaking correspondence with the current scene. Across all four auxiliary sources, matched observations consistently yield better reconstruction and reference-based structural agreement. We further perform patch-level interventions with fixed initial noise, using the resulting reference-free output change to rank spatial evidence. Replacing top-ranked patches causes substantially larger PSNR and Edge-F1 degradation than random replacement for NIR, SAR, and temporal evidence, providing a direct faithfulness test of the spatial attribution. Mechanistic controls further show that freezing the shared backbone or detaching condition--generation interaction weakens reconstruction, supporting a functional role for the evidence pathway.

Finally, progressively increasing heterogeneous evidence across multichannel, multimodal, and multitemporal settings consistently reduces both prediction-only and missing structures. Thus, the additional observations do more than improve aggregate restoration metrics: they progressively constrain the degrees of freedom of the pretrained generative prior. Nevertheless, residual reference-inconsistent structures remain under severe ambiguity, indicating that evidence grounding mitigates but does not guarantee reliable reconstruction.

Our contributions are summarized as follows:
\begin{itemize}
    \item We formulate cloud removal as conditional generation from heterogeneous remote sensing observations and introduce \textbf{\method}, which combines the completion capability of a strong pretrained generative prior with explicit source-specific condition-token blocks for observational evidence.

    \item We characterize a reliability trade-off of strong-prior restoration and perform source-level interventional attribution. Structural auditing reveals reduced omission but increased prediction-only completion relative to specialist baselines, while scene-correspondence interventions show consistent functional dependence on matched NIR, spectral, SAR, and temporal evidence.

    \item We perform spatial attribution and rigorous faithfulness evaluation through controlled patch replacement. Top-attributed evidence causes substantially greater degradation than random intervention, while progressively richer heterogeneous evidence reduces structural inconsistency, showing that observational evidence provides an effective but incomplete constraint on prior-driven generation.
\end{itemize}

\section{Related Work} \label{sec:related} \paragraph{Remote Sensing Cloud Removal.} Remote-sensing cloud removal is a long-standing restoration problem with several common settings, including single-image restoration, multimodal restoration, and multitemporal restoration \cite{cloudremovalsurvey2025jstars}. Early work often focused on recovering clean content from optical imagery alone or from optical inputs with auxiliary spectral guidance \cite{filmycloud2017cvprw,huang2022ctgan}. Subsequent studies extended the problem to multimodal settings by incorporating complementary SAR observations for optical reconstruction \cite{meraner2020cloud,sen12mscr2020tgrs,glfcr2022isprs,ussrn2024tgrs}. Another major line of work exploits temporal redundancy, ranging from classical matrix-completion formulations such as TECROMAC \cite{tecromac2016cvpr} to learned multitemporal restoration models such as STGAN \cite{stgan2020wacv} and UnCRtainTS \cite{ebel2023uncrtaints}. More recent generative approaches, including diffusion-based models, further improve restoration quality within specific settings \cite{zou2024diffcr,liu2025effective}. Overall, however, these regimes are still mostly handled as separate problem variants with different backbones, fusion mechanisms, and dataset-specific designs. Rather than using richer observations only to improve restoration,
\method{} preserves them as explicit evidence sources and analyzes how
they constrain a strong generative prior under controlled intervention.

\paragraph{Unified Multimodal Models.}
Unified multimodal models aim to organize understanding and generation within a shared framework. 
Existing approaches include autoregressive visual token modeling~\cite{janus2024,januspro2025,emu3,lu2024unified,vila-u,wu2024liquid,team2024chameleon,xie2024muse,qu2024tokenflow}, modular systems that connect multimodal backbones to separate generative interfaces~\cite{emu2,seed-x,dongdreamllm,pan2025transfer,tong2024metamorph}, and integrated transformer designs that more tightly couple understanding with diffusion or flow-style generation~\cite{transfusion,janusflow2024,shi2024llamafusion,liang2024mixture}. 
The last direction is closest to our setting because condition tokens and generation tokens can interact within a shared backbone. 
In particular, BAGEL~\cite{deng2025emerging} organizes visual condition tokens and latent generation tokens in a shared token-based framework. 
\method{} adapts this token-based interface to remote sensing while
preserving source and spatial organization, enabling source-wise and
patch-level analysis of evidence dependence and attribution faithfulness.

\section{Method}
\label{sec:method}

\subsection{Overview}
\label{subsec:overview}

\begin{figure*}[t]
    \centering
    \includegraphics[width=\linewidth]{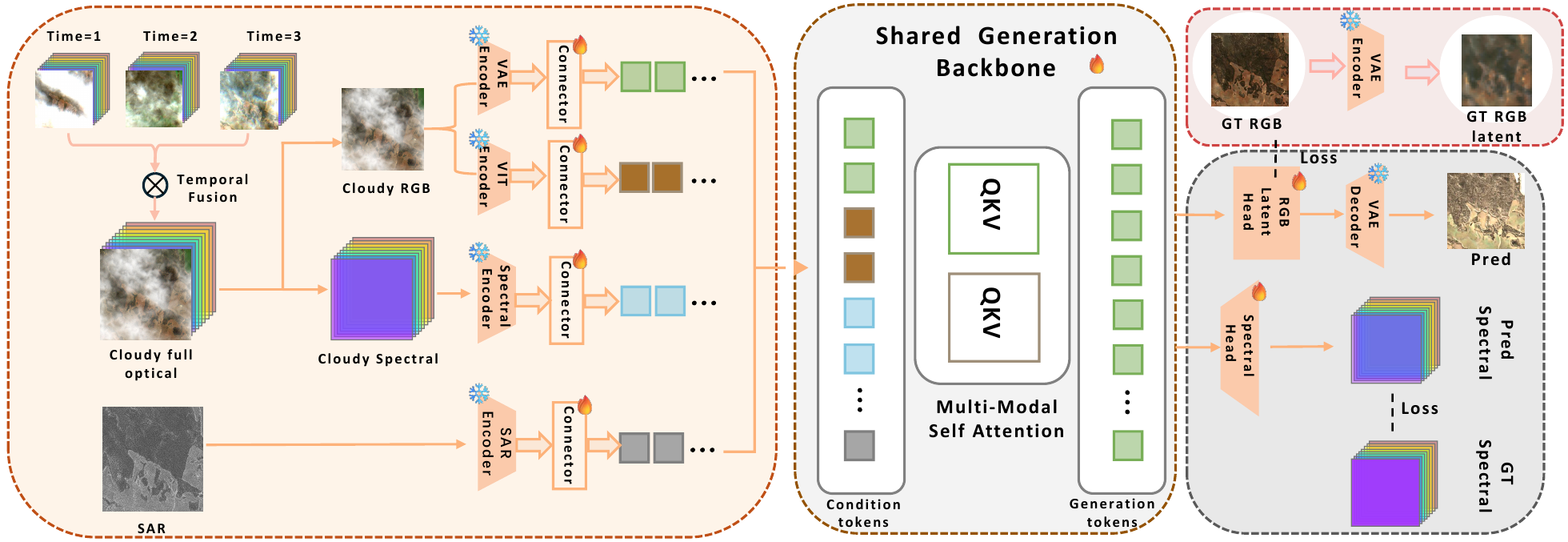}
    \caption{
    Overview of \method{}.
    Source-specific encoders and connectors map heterogeneous observations
    into packed condition-token blocks, which guide RGB latent generation
    and optional non-RGB spectral prediction through a shared pretrained
    generation backbone.
    }
    \label{fig:unicr_framework}
\end{figure*}

We propose \method{}, a generative cloud-removal framework for heterogeneous
remote sensing observations. As shown in Fig.~\ref{fig:unicr_framework},
source-specific encoders and connectors project available observations into
condition tokens, which interact with noisy RGB latent tokens in a shared
pretrained BAGEL backbone~\cite{deng2025emerging}.

For a scene,
\(\mathcal{X}=\{\mathbf{o}^{i}\}_{i\in\mathcal{A}}\)
denotes the available RGB, non-RGB optical, SAR, or temporal observations.
The target contains RGB
\(\mathbf{x}^{\mathrm{rgb}}_0\) and, when available, spectral bands
\(\mathbf{x}^{\mathrm{band}}_0\).
RGB uses the BAGEL visual branch, while spectral and SAR observations use
frozen SpectralGPT~\cite{hong2024spectralgpt} and
SARMAE~\cite{liu2025sarmae} encoders, respectively. The same backbone and
condition-token interface are reused across multichannel, multimodal, and
multitemporal configurations.

\subsection{Heterogeneous Evidence Interface}
\label{subsec:evidence_interface}

For multitemporal optical observations, \method{} applies a deterministic
reliability mechanism before tokenization. At each location \(p\), the gate
uses cloud-related cues, including brightness, whiteness, and temporal
disagreement, to assign normalized weights \(w_l(p)\) to the available
frames. The fused optical observation is

\begin{equation}
    \bar{\mathbf{x}}^{\mathrm{opt}}(p)
    =
    \sum_{l=1}^{L}
    w_l(p)\mathbf{x}^{\mathrm{opt}}_l(p),
    \qquad
    \sum_{l=1}^{L}w_l(p)=1.
    \label{eq:compact_temporal_fusion}
\end{equation}

For a single temporal observation, the fusion reduces to the identity.
SAR evidence is not passed through the optical temporal gate.
The reliability weights provide an explicit input-level indication of
which temporal observations are emphasized at each location. Additional
matched controls of the reliability mechanism are reported in
Appendix~\ref{app:temporal_controls}.

Each available source is then projected into the shared hidden dimension.
For source \(i\in\mathcal{A}\), the condition-token block and the complete
condition sequence are

\begin{equation}
\begin{aligned}
    \mathbf{T}^{i}
    &=
    P_i\!\left(E_i(\mathbf{o}^{i})\right)
    +
    \mathbf{e}^{i}_{\mathrm{pos}},
    \\
    \mathbf{T}_{\mathcal{X}}
    &=
    \operatorname{Pack}
    \left(
    \{\mathbf{T}^{i}\}_{i\in\mathcal{A}}
    \right),
\end{aligned}
\label{eq:compact_condition_tokens}
\end{equation}

where \(E_i\) and \(P_i\) denote the source-specific encoder and connector,
respectively, and \(\mathbf{e}^{i}_{\mathrm{pos}}\) denotes the positional
encoding of the corresponding token block. Heterogeneous observations are
therefore aligned in the shared representation space without being forced
into a single pixel-level tensor.

Because each source occupies a distinct token block and its positional
encoding preserves within-source spatial organization, the packed
representation supports both source-level replacement and patch-level
intervention. These properties enable the evidence-dependence and
faithfulness analyses in Secs.~\ref{subsec:evidence_grounding}
and~\ref{subsec:spatial_faithfulness}.

\subsection{Conditional Generation and Prediction}
\label{subsec:generation_objective}

The clean RGB target is encoded by the frozen BAGEL VAE into latent tokens
\(\mathbf{z}_0\). Following flow matching, we sample
\(\boldsymbol{\epsilon}\sim\mathcal{N}(\mathbf{0},\mathbf{I})\) and
\(t\in[0,1]\), and construct

\begin{equation}
    \mathbf{z}_t
    =
    (1-t)\mathbf{z}_0+t\boldsymbol{\epsilon},
    \qquad
    \mathbf{v}^{\star}
    =
    \boldsymbol{\epsilon}-\mathbf{z}_0.
    \label{eq:compact_flow_matching}
\end{equation}

The noisy latent is projected into generation tokens and packed with the
condition sequence:

\begin{equation}
\begin{aligned}
    \mathbf{G}_t
    &=
    \Phi_{\mathrm{in}}(\mathbf{z}_t)
    +
    \mathbf{e}^{t}(t)
    +
    \mathbf{e}^{\mathrm{g}}_{\mathrm{pos}},
    \\
    \mathbf{Y}_{\mathrm{g}}
    &=
    \operatorname{Select}_{\mathrm{g}}
    \left[
    F_{\theta}
    \left(
    \operatorname{Pack}
    \left(
    \mathbf{T}_{\mathcal{X}},
    \mathbf{G}_t
    \right)
    \right)
    \right],
\end{aligned}
\label{eq:compact_generation}
\end{equation}

where \(F_{\theta}\) is the pretrained BAGEL shared generation backbone and
\(\mathbf{Y}_{\mathrm{g}}\) denotes the conditioned generation states.
An RGB latent head predicts the flow velocity from
\(\mathbf{Y}_{\mathrm{g}}\). When non-RGB supervision is available, a
lightweight spectral head predicts the remaining optical bands from the same
conditioned states.

The overall objective is

\begin{equation}
    \mathcal{L}_{\mathrm{UniCR}}
    =
    \mathcal{L}_{\mathrm{rgb}}
    +
    \lambda_{\mathrm{band}}
    \mathbb{I}_{\mathrm{band}}
    \mathcal{L}_{\mathrm{band}},
    \label{eq:compact_final_loss}
\end{equation}

where \(\mathcal{L}_{\mathrm{rgb}}\) is the RGB velocity-matching loss,
\(\mathcal{L}_{\mathrm{band}}\) is the non-RGB spectral prediction loss, and
\(\mathbb{I}_{\mathrm{band}}\) indicates whether spectral supervision is
available. During inference, the model starts from a Gaussian RGB latent,
solves the learned velocity field, and decodes the resulting latent through
the frozen VAE. Predicted spectral patches are merged with the RGB output
when full multispectral reconstruction is required.

\section{Experiments}
\label{sec:experiments}

We evaluate \method{} on CUHK-CR1/2~\cite{sui2024diffusion},
SEN12MS-CR~\cite{sen12mscr2020tgrs}, and
Sen2\_MTC\_New~\cite{huang2022ctgan}, covering multichannel,
multimodal, and multitemporal cloud removal.

\subsection{Experimental Setup}
\label{subsec:experimental_setup}

\method{} uses the pretrained BAGEL backbone~\cite{deng2025emerging}.
Observation encoders and the VAE are frozen, while the connectors, shared
backbone, and output heads are optimized on two NVIDIA GH200 GPUs.
Implementation details are provided in Appendix~\ref{app:experimental_details}.

We use standard benchmark metrics together with prediction-only structures,
missing structures, and Edge F1 for reference-based structural analysis.
Source-wise attribution replaces matched auxiliary evidence with
same-modality evidence from another scene while fixing the RGB anchor and
initial noise. Spatial evidence groups are ranked by their reference-free
output effect, and attribution faithfulness is evaluated by top-ranked versus
random replacement using AOPC. We report paired-bootstrap confidence
intervals, with ROI-cluster bootstrap for SEN12MS-CR. Full protocols are in
Appendix~\ref{app:analysis_protocols}.

\subsection{Restoration Performance and Failure Profile}
\label{subsec:performance}

\begin{figure*}[t]
    \centering
    \includegraphics[width=\linewidth]{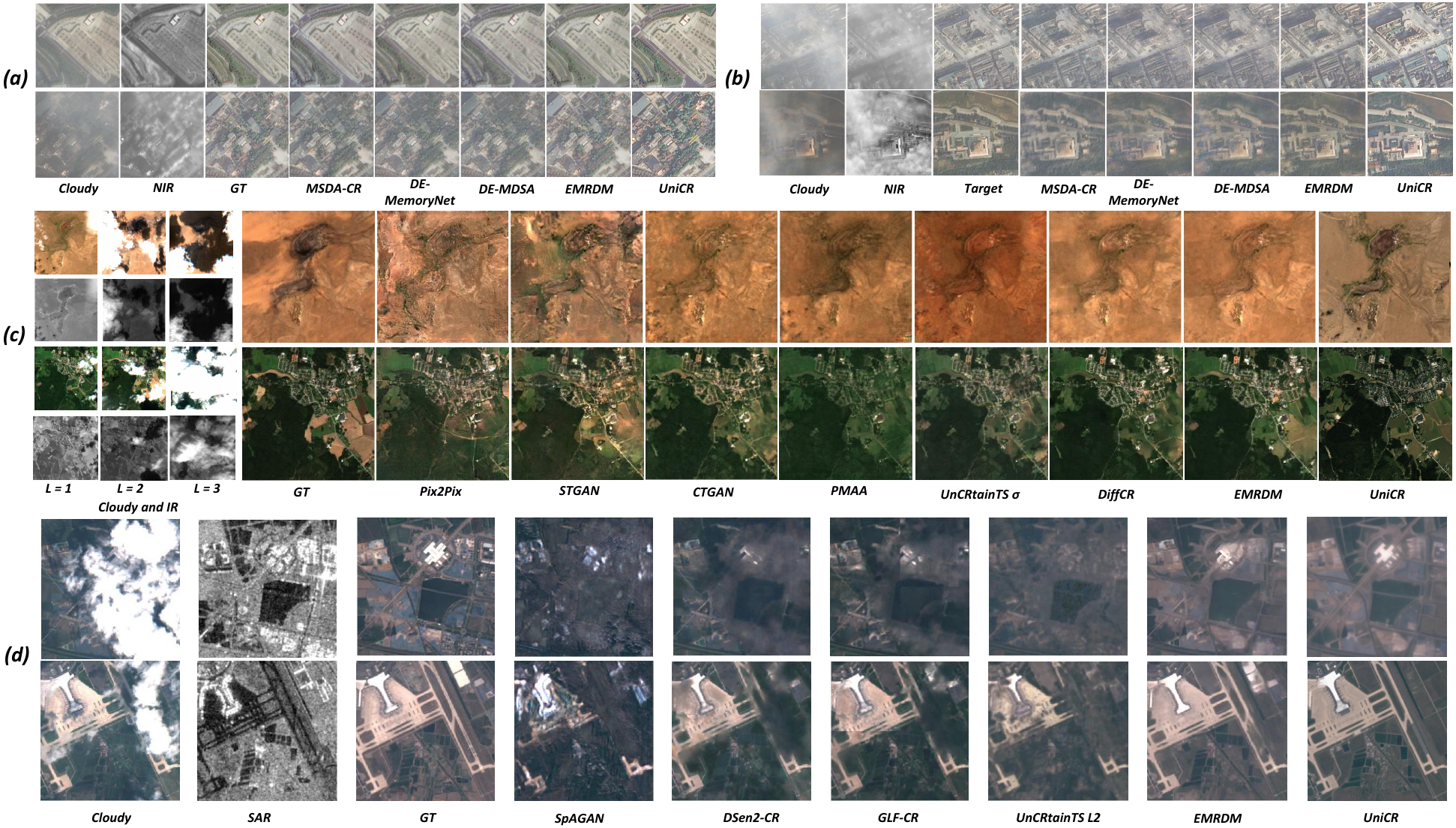}
    \caption{Qualitative comparison across the four evaluated benchmarks.}
    \label{fig:main_results}
\end{figure*}

\begin{table}[t]
\centering
\caption{
Comparison with representative cloud-removal methods.
Best and second-best results are shown in \textbf{bold} and underlined,
respectively.
}
\label{tab:main_results}

\small
\setlength{\tabcolsep}{5pt}

\textbf{(a) CUHK-CR1 and CUHK-CR2}
\vspace{0.25em}

\resizebox{0.88\textwidth}{!}{
\begin{tabular}{lcccccc}
\toprule
\multirow{2}{*}{Method}
&
\multicolumn{3}{c}{CUHK-CR1}
&
\multicolumn{3}{c}{CUHK-CR2}
\\
\cmidrule(lr){2-4}
\cmidrule(lr){5-7}
&
PSNR$\uparrow$
&
SSIM$\uparrow$
&
LPIPS$\downarrow$
&
PSNR$\uparrow$
&
SSIM$\uparrow$
&
LPIPS$\downarrow$
\\
\midrule
SpA-GAN~\cite{emami2020spa} & 20.999 & 0.5162 & 0.0830 & 19.680 & 0.3952 & 0.1201 \\
AMGAN-CR~\cite{xu2022attention} & 20.867 & 0.4986 & 0.1075 & 20.172 & 0.4900 & 0.0930 \\
CVAE~\cite{bao2017cvae} & 24.252 & 0.7252 & 0.1075 & 22.631 & 0.6302 & 0.0489 \\
MemoryNet~\cite{zhang2026memory} & 26.073 & 0.7741 & 0.0315 & 24.224 & 0.6838 & 0.0403 \\
MSDA-CR~\cite{yu2022cloud} & 25.435 & 0.7483 & 0.0374 & 23.755 & 0.6661 & 0.0433 \\
DE-MemoryNet~\cite{sui2024diffusion} & 26.183 & 0.7746 & 0.0290 & 24.348 & 0.6843 & 0.0369 \\
DE-MSDA~\cite{sui2024diffusion} & 25.739 & 0.7592 & 0.0321 & 23.968 & 0.6737 & 0.0372 \\
EMRDM~\cite{liu2025effective} & \underline{27.281} & \underline{0.8007} & \underline{0.0218} & \underline{24.594} & \underline{0.6951} & \underline{0.0301} \\
\midrule
\method{} & \textbf{28.310} & \textbf{0.8170} & \textbf{0.0189} & \textbf{25.610} & \textbf{0.7100} & \textbf{0.0272} \\
\bottomrule
\end{tabular}
}

\vspace{0.8em}

\noindent
\hspace*{-0.015\textwidth}
\begin{minipage}[t]{0.47\textwidth}
\centering
\vspace{0pt}
\textbf{(b) Sen2\_MTC\_New}
\vspace{0.25em}

\setlength{\tabcolsep}{3.2pt}
\begin{tabular}{lccc}
\toprule
Method
&
PSNR$\uparrow$
&
SSIM$\uparrow$
&
LPIPS$\downarrow$
\\
\midrule
Pix2Pix~\cite{henry2021pix2pix} & 16.985 & 0.455 & 0.535 \\
DSen2-CR~\cite{meraner2020cloud} & 16.827 & 0.534 & 0.446 \\
STGAN~\cite{stgan2020wacv} & 18.152 & 0.587 & 0.513 \\
CTGAN~\cite{huang2022ctgan} & 18.308 & 0.609 & 0.384 \\
PMAA~\cite{zou2023pmaa} & 18.369 & 0.614 & 0.392 \\
UnCRtainTS~\cite{ebel2023uncrtaints} & 18.770 & 0.631 & 0.333 \\
DDPM-CR~\cite{jing2023denoising} & 18.742 & 0.614 & 0.329 \\
DiffCR~\cite{zou2024diffcr} & 19.150 & 0.671 & 0.291 \\
EMRDM~\cite{liu2025effective} & \underline{20.067} & \underline{0.709} & \underline{0.255} \\
\midrule
\method{} & \textbf{21.050} & \textbf{0.724} & \textbf{0.230} \\
\bottomrule
\end{tabular}
\end{minipage}
\hspace{0.018\textwidth}
\begin{minipage}[t]{0.50\textwidth}
\centering
\vspace{0pt}
\textbf{(c) SEN12MS-CR}
\vspace{0.25em}

\setlength{\tabcolsep}{2.2pt}
\begin{tabular}{lcccc}
\toprule
Method
&
PSNR$\uparrow$
&
SSIM$\uparrow$
&
MAE$\downarrow$
&
SAM$\downarrow$
\\
\midrule
SAR-Opt-cGAN~\cite{grohnfeldt2018conditional} & 25.59 & 0.764 & 0.043 & 15.494 \\
SAR2OPT~\cite{toriya2019sar2opt} & 25.87 & 0.793 & 0.042 & 14.788 \\
SpA-GAN~\cite{emami2020spa} & 24.78 & 0.754 & 0.045 & 18.085 \\
DSen2-CR~\cite{meraner2020cloud} & 27.76 & 0.874 & 0.031 & 9.472 \\
GLF-CR~\cite{xu2022glf} & 28.64 & 0.885 & 0.028 & 8.981 \\
UnCRtainTS L2~\cite{ebel2023uncrtaints} & 28.90 & 0.880 & 0.027 & 8.320 \\
ACA-Net~\cite{huang2024attentive} & 29.78 & 0.896 & 0.025 & 7.770 \\
DiffCR~\cite{zou2024diffcr} & 31.77 & 0.902 & 0.019 & 5.821 \\
EMRDM~\cite{liu2025effective} & \underline{32.14} & \underline{0.924} & \underline{0.018} & \underline{5.267} \\
\midrule
\method{} & \textbf{33.120} & \textbf{0.934} & \textbf{0.016} & \textbf{4.900} \\
\bottomrule
\end{tabular}
\end{minipage}
\end{table}

Table~\ref{tab:main_results} confirms strong restoration performance across
all four settings, while Fig.~\ref{fig:main_results} shows sharp and coherent
completion under severe cloud contamination. We next examine whether this
strong prior changes the structural failure profile.

Table~\ref{tab:failure_mechanistic}(a) audits the same 111 CUHK-CR2 outputs.
Compared with specialist baselines, \method{} has fewer missing structures
but more prediction-only structures. Thus, the strong prior shifts the
failure balance from conservative omission toward unsupported completion
rather than uniformly reducing all errors. Because the compared systems
also differ in architecture and training, we interpret this as an observed
failure-profile difference rather than an effect of model scale alone.

\begin{table}[t]
\centering
\caption{
Complementary analysis on CUHK-CR2.
(a) Structural failure-profile audit.
(b) Conditioning, adaptation, and condition--generation pathway controls.
}
\label{tab:failure_mechanistic}
\scriptsize
\setlength{\tabcolsep}{4pt}

\begin{minipage}[t]{0.43\textwidth}
\centering
\textbf{(a) Structural failure profile}
\vspace{0.1em}

\begin{tabular}{lccc}
\toprule
Method
& Pred.-only$\downarrow$
& Missing$\downarrow$
& Edge F1$\uparrow$ \\
\midrule
MemoryNet~\cite{zhang2026memory}
& 0.210 & 0.465 & 0.638 \\
EMRDM~\cite{liu2025effective}
& 0.235 & 0.405 & 0.671 \\
DiffCR~\cite{zou2024diffcr}
& 0.260 & 0.385 & 0.671 \\
\method{}
& 0.323 & \textbf{0.352} & 0.662 \\
\bottomrule
\end{tabular}
\end{minipage}
\hfill
\begin{minipage}[t]{0.55\textwidth}
\centering
\textbf{(b) Conditioning and pathway controls}
\vspace{0.1em}

\resizebox{\linewidth}{!}{
\begin{tabular}{lccc}
\toprule
Variant
& PSNR$\uparrow$
& SSIM$\uparrow$
& LPIPS$\downarrow$ \\
\midrule
BAGEL; no CR training
& 14.60 & 0.300 & 0.145 \\
Native RGB full FT
& 19.18 & 0.513 & 0.080 \\
Frozen shared backbone
& 21.90 & 0.610 & 0.064 \\
Packed RGB-only
& 23.997 & 0.6629 & 0.0340 \\
Detached condition--generation pathway
& 24.40 & 0.680 & 0.037 \\
Full RGB+NIR \method{}
& \textbf{25.610}
& \textbf{0.7100}
& \textbf{0.0272} \\
\bottomrule
\end{tabular}
}
\end{minipage}

\end{table}

\subsection{Interventional and Mechanistic Evidence Analysis}
\label{subsec:evidence_grounding}

Table~\ref{tab:failure_mechanistic}(b) shows that direct BAGEL use, native
RGB fine-tuning, a frozen backbone, and detached condition--generation
interaction all underperform the full model. Thus, strong reconstruction
requires not only the pretrained prior, but also task adaptation and
functional interaction with condition evidence.

Source-wise interventions in Table~\ref{tab:interventional_analysis}(a)
provide a stricter test. Replacing matched observations with foreign-scene
evidence consistently degrades both PSNR and Edge F1 for NIR, spectral,
SAR, and temporal sources, with all bootstrap intervals excluding zero.
\method{} therefore depends on scene-corresponding evidence rather than
token presence alone. The same-modality replacement preserves modality type
and token count and is less distribution-disruptive than zero masking.
Because sources are evaluated in different regimes, their effect sizes are
not interpreted as a universal ranking of modality importance.

\begin{table}[!t]
\centering
\caption{
Interventional analysis of heterogeneous evidence.
(a) Source-wise scene-correspondence interventions.
(b) Faithfulness of intervention-derived spatial attribution.
For lower-is-better LPIPS and SAM, the secondary column reports the
matched-to-shuffled change directly.
}
\label{tab:interventional_analysis}
\scriptsize
\setlength{\tabcolsep}{3.2pt}

\textbf{(a) Source-wise scene-correspondence intervention}
\vspace{0.1em}

\resizebox{\textwidth}{!}{
\begin{tabular}{llccccc}
\toprule
Evidence
& Regime
& PSNR: matched$\rightarrow$shuffled
& $\Delta$PSNR [95\% CI]
& Secondary: matched$\rightarrow$shuffled
& Edge F1: matched$\rightarrow$shuffled
& $\Delta$F1 [95\% CI] \\
\midrule
NIR
& CUHK-CR2
& $25.61\rightarrow24.08$
& $+1.53\ [1.12,1.94]$
& LPIPS $0.0272\rightarrow0.0415$
& $0.662\rightarrow0.594$
& $+0.068\ [0.047,0.089]$ \\
Spectral
& SEN12MS-CR
& $33.12\rightarrow31.76$
& $+1.36\ [0.98,1.75]$
& SAM $4.90\rightarrow5.72$
& $0.674\rightarrow0.631$
& $+0.043\ [0.028,0.058]$ \\
SAR
& SEN12MS-CR
& $33.12\rightarrow30.91$
& $+2.21\ [1.73,2.69]$
& SAM $4.90\rightarrow6.18$
& $0.674\rightarrow0.616$
& $+0.058\ [0.039,0.077]$ \\
Temporal
& Sen2\_MTC\_New
& $21.10\rightarrow18.96$
& $+2.14\ [1.70,2.58]$
& LPIPS $0.229\rightarrow0.256$
& $0.660\rightarrow0.603$
& $+0.057\ [0.038,0.076]$ \\
\bottomrule
\end{tabular}
}

\vspace{0.35em}

\textbf{(b) Top-ranked versus random spatial intervention}
\vspace{0.1em}

\resizebox{0.88\textwidth}{!}{
\begin{tabular}{lcccccc}
\toprule
Source
& \multicolumn{3}{c}{PSNR AOPC}
& \multicolumn{3}{c}{Edge-F1 AOPC} \\
\cmidrule(lr){2-4}
\cmidrule(lr){5-7}
& Top
& Random
& Gap [95\% CI]
& Top
& Random
& Gap [95\% CI] \\
\midrule
NIR
& 2.05 & 0.74
& $1.31\ [1.01,1.62]$
& 0.069 & 0.024
& $0.045\ [0.031,0.059]$ \\
SAR
& 2.88 & 0.98
& $1.90\ [1.47,2.33]$
& 0.081 & 0.028
& $0.052\ [0.036,0.069]$ \\
Temporal
& 1.89 & 0.68
& $1.21\ [0.92,1.51]$
& 0.066 & 0.023
& $0.043\ [0.029,0.057]$ \\
\bottomrule
\end{tabular}
}

\end{table}

\begin{figure}[!t]
    \centering
    \includegraphics[width=\linewidth]{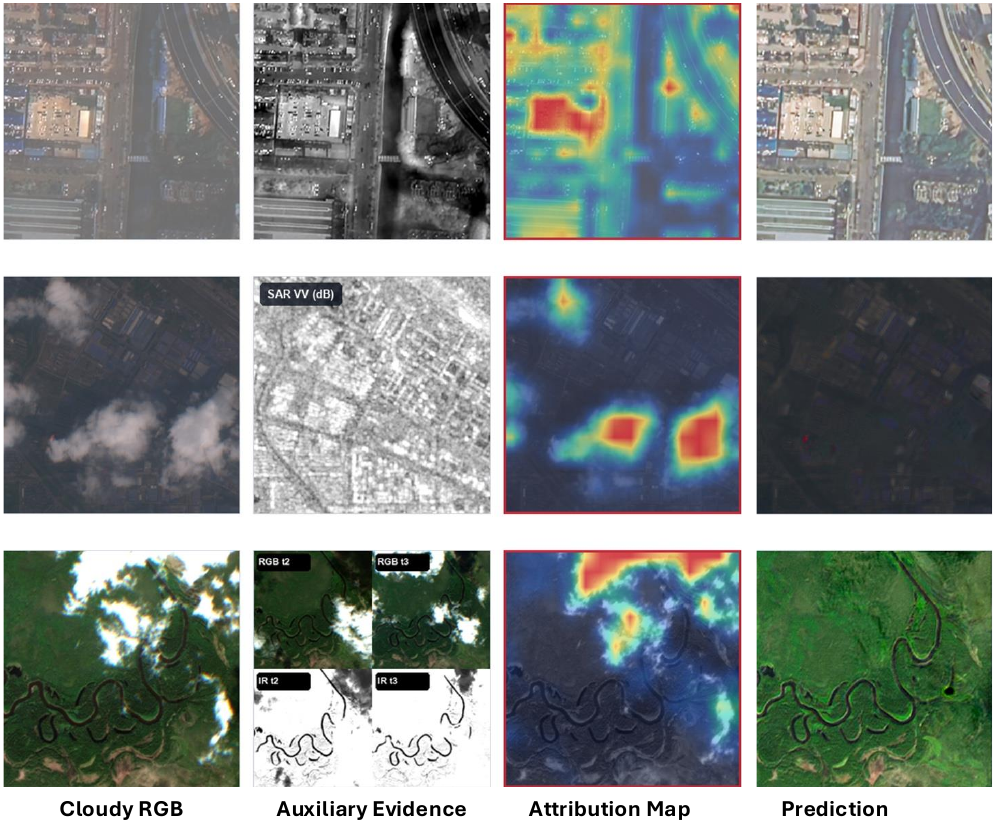}
    \caption{
     Representative spatial-attribution examples for NIR (top), SAR (middle),
    and temporal evidence (bottom). From left to right: cloudy RGB input,
    auxiliary evidence, intervention-derived attribution map, and final
    prediction. Warmer colors indicate larger output changes under
    patch-level foreign-scene replacement and therefore stronger functional
    dependence on the corresponding auxiliary evidence.
    }
    \label{fig:spatial_attr_examples}
\end{figure}
\subsection{Spatial Attribution and Faithfulness}
\label{subsec:spatial_faithfulness}

Figure~\ref{fig:spatial_attr_examples} shows intervention-derived spatial
attribution for representative NIR, SAR, and temporal evidence, highlighting
where auxiliary observations most strongly constrain reconstruction under
optical ambiguity.

Table~\ref{tab:interventional_analysis}(b) validates the ranking without
using the clean reference to construct it: replacing top-attributed groups
causes substantially larger PSNR and Edge-F1 degradation than random
replacement, with all AOPC gaps positive and bootstrap intervals excluding
zero. This supports attribution faithfulness, while the analysis remains
limited to representative sources without semantic interpretation of the
selected patches.

\begin{figure}[!t]
    \centering
    \includegraphics[width=\linewidth]{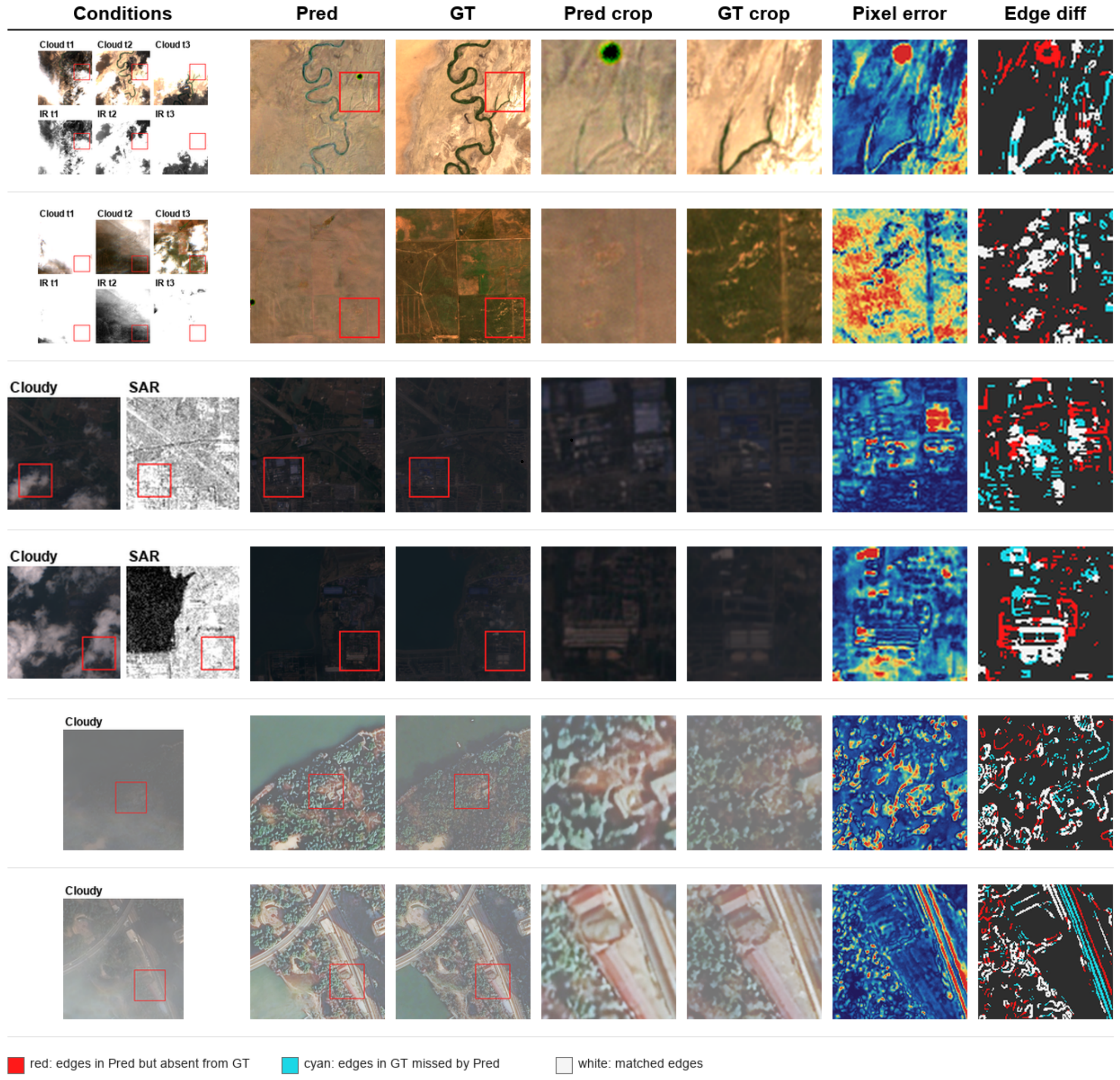}
    \caption{
    Residual failures across heterogeneous settings.
    Red/cyan/white denote prediction-only, missing, and matched edges.
    }
    \label{fig:structural_analysis}
\end{figure}

\subsection{Evidence Sufficiency and Residual Failure Analysis}
\label{subsec:evidence_sufficiency}

Richer evidence consistently improves structural agreement. Edge F1 rises
from \(0.610\) to \(0.662\) with NIR on CUHK-CR2, from \(0.580\) to
\(0.626/0.674\) with spectral/SAR evidence on SEN12MS-CR, and from \(0.556\)
to \(0.660\) as temporal context increases from \(L=1\) to \(L=3\).
Prediction-only and missing structures decrease correspondingly, indicating
that additional scene evidence constrains both omission and unsupported
completion. Full ablations are in
Appendix~\ref{app:full_condition_ablations}.

Residual failures remain under severe ambiguity
(Fig.~\ref{fig:structural_analysis}), where plausible boundaries or textures
may be absent or shifted relative to the paired reference. Heterogeneous
evidence therefore mitigates, but does not eliminate, prior-driven
structural inconsistency or guarantee physical correctness.
\FloatBarrier

\section{Conclusion}
\label{sec:conclusion}

We presented \method{}, whose source-explicit condition interface enables interventional analysis of heterogeneous evidence in generative cloud removal. Structural auditing shows that a strong prior reduces missing structures but increases prediction-only completion relative to specialist baselines. Scene-correspondence and patch-level interventions demonstrate functional dependence on matched evidence and validate attribution faithfulness. Richer multichannel, multimodal, and multitemporal observations further reduce structural inconsistency, although residual reference-inconsistent content remains. Heterogeneous evidence therefore constrains, but cannot guarantee the correctness of, prior-driven reconstruction.

\bibliographystyle{plain}
\bibliography{references}

\newpage

\appendix

\section*{Technical Appendices and Supplementary Material}

%%%%%%%%%%%%%%%%%%%%%%%%%%%%%%%%%%%%%%%%%%%%%%%%%%%%%%%%%%%%
\section{Additional Experimental Details}
\label{app:experimental_details}

\subsection{Datasets}
\label{app:datasets}

We conduct experiments on four representative cloud-removal benchmarks:
CUHK-CR1, CUHK-CR2, SEN12MS-CR, and Sen2\_MTC\_New. These datasets cover
different observation regimes, including RGB images with aligned auxiliary
optical bands, multispectral optical images with SAR measurements, and
multitemporal optical observations.

The CUHK-CR1 and CUHK-CR2 datasets contain high-resolution satellite images
with RGB and near-infrared (NIR) bands. CUHK-CR1 mainly contains thin-cloud
scenes, while CUHK-CR2 contains thicker cloud contamination. Following the
standard protocol, CUHK-CR1 is split into 534 training images and 134 testing
images, and CUHK-CR2 is split into 448 training images and 111 testing images.
RGB is treated as the main generation target, while NIR is used as aligned
auxiliary optical evidence.

The SEN12MS-CR dataset contains co-registered Sentinel-2 multispectral optical
observations and Sentinel-1 SAR measurements. The optical image has 13 bands,
and the SAR image has two channels. We follow the commonly used split with
114,050 training samples, 7,176 validation samples, and 7,899 testing samples.
For this dataset, \method{} restores the full cloud-free optical target by
decoding RGB from the latent generation branch and predicting the remaining
non-RGB optical bands through the spectral head.

The Sen2\_MTC\_New dataset contains temporally ordered optical observations.
Each temporal frame contains RGB channels and an IR channel. Following prior
work, the dataset contains 2,380 training samples, 350 validation samples, and
687 testing samples. Unless otherwise specified, we use a temporal context
length of \(L=3\). The temporal observations are fused by the deterministic
reliability mechanism described in Sec.~\ref{subsec:evidence_interface}, and the
model restores the RGB target.

\subsection{Pre-processing}
\label{app:preprocessing}

For CUHK-CR1 and CUHK-CR2, the original images are resized from
\(512\times512\) to \(256\times256\). Pixel values are converted from integer
values in \([0,255]\) to floating-point values in \([0,1]\). RGB images are
used for the generation target and RGB condition branch, while the aligned NIR
band is resized with the same spatial transform and used as auxiliary optical
evidence.

For SEN12MS-CR, optical values are clipped to \([0,10000]\) and rescaled to
\([0,1]\). SAR values are clipped to a fixed backscatter range and normalized
before being passed to the SAR encoder. The RGB subset of the optical tensor is
used by the RGB condition and generation branches, while the remaining non-RGB
optical bands are supervised through the spectral branch. SAR is used only as
condition evidence and is not reconstructed.

For Sen2\_MTC\_New, all optical values are divided by \(10000\) and clipped to
\([0,1]\). Each temporal frame is spatially aligned and resized to
\(256\times256\). The temporal reliability gate is applied to the full optical
tensor before tokenization. After temporal fusion, RGB is used as the main
optical condition and IR as auxiliary optical evidence.

\subsection{Token Alignment}
\label{app:token_alignment}

\method{} aligns heterogeneous observations at the token level. Different
observation sources are processed by their corresponding frozen encoders and
projected into the shared hidden dimension by trainable connectors. The
resulting source-specific token blocks are concatenated into a packed condition
sequence.

For RGB visual conditions, the BAGEL visual branch produces
\(18\times18\) condition tokens. For RGB generation, the clean target is
encoded by the frozen VAE and patchified into \(16\times16\) latent generation
tokens.

For spectral optical observations, the frozen SpectralGPT encoder produces
spectral-spatial tokens with shape \(4\times16\times16\). We apply group-wise
\(2\times2\) average pooling, resulting in \(4\times8\times8=256\) spectral
tokens before projection by the spectral connector.

For SAR observations, the frozen SARMAE encoder produces patch tokens at a
\(14\times14\) spatial resolution after removing the class token, resulting
in 196 SAR condition tokens. No SAR reconstruction objective is introduced.

\subsection{Training Configuration}
\label{app:training_configuration}

The pretrained observation encoders and VAE remain frozen throughout training.
The modality-specific connectors, shared generation backbone, RGB latent head,
and spectral head are optimized for cloud removal. All benchmarks use the same
optimization recipe unless otherwise specified.

\begin{table}[t]
\centering
\caption{Common training configuration of \method{}.}
\label{tab:common_training_recipe}
\small
\setlength{\tabcolsep}{5pt}
\resizebox{0.78\linewidth}{!}{
\begin{tabular}{ll}
\toprule
Item & Setting \\
\midrule
Backbone initialization & BAGEL-7B-MoT \\
Frozen modules & VAE and observation encoders \\
Trainable modules & Connectors, backbone, output heads \\
Precision & bfloat16 \\
Parallelism & FSDP on 2$\times$ GH200 \\
Optimizer & AdamW \\
AdamW betas & $(0.9,0.95)$ \\
Weight decay & $0$ \\
Learning rate & \(1.0\times10^{-5}\) \\
LR schedule & Cosine decay \\
Warm-up & 500 steps \\
Per-device batch size & 1 \\
Gradient accumulation & 1 \\
Token budget & \(\sim\)28K tokens per rank \\
\bottomrule
\end{tabular}
}
\end{table}

Table~\ref{tab:benchmark_training_setup} summarizes the dataset-specific
condition sources and prediction targets. The same architecture and
condition-token contract are reused across these settings, while separate
checkpoints are adapted to their corresponding sensor configurations.

\begin{table}[t]
\centering
\caption{Benchmark-specific condition sources and prediction targets.}
\label{tab:benchmark_training_setup}
\small
\setlength{\tabcolsep}{4.5pt}
\resizebox{0.94\linewidth}{!}{
\begin{tabular}{llll}
\toprule
Benchmark & Condition sources & Output & Checkpoint \\
\midrule
CUHK-CR1
& RGB + NIR
& RGB
& 6K \\
CUHK-CR2
& RGB + NIR
& RGB
& 6K \\
SEN12MS-CR
& RGB + spectral + SAR
& 13-band optical
& 6K \\
Sen2\_MTC\_New
& temporal RGB + IR
& RGB
& 6K \\
\bottomrule
\end{tabular}
}
\end{table}

\subsection{Evaluation Metrics}
\label{app:evaluation_metrics}

We follow the standard evaluation protocol of each benchmark. For CUHK-CR1
and CUHK-CR2, restored RGB images are evaluated using PSNR, SSIM, and LPIPS.
LPIPS is computed using the standard pretrained AlexNet backbone.

For SEN12MS-CR, the RGB prediction and non-RGB spectral prediction are merged
into the original 13-band optical order. PSNR, SSIM, MAE, and spectral angle
mapper (SAM) are computed on the reconstructed multispectral output.

For Sen2\_MTC\_New, PSNR, SSIM, and LPIPS are computed on the restored RGB
target.

In addition to these standard metrics, the main paper reports prediction-only
structures, missing structures, and Edge F1 as reference-based structural
diagnostics. These measures quantify structural disagreement with the paired
clean target and should not be interpreted as direct validation of physical
correctness.

\section{Evaluation and Intervention Protocols} \label{app:analysis_protocols} \subsection{Structural Failure Audit} \label{app:structural_audit} The CUHK-CR2 failure-profile audit is conducted on the same 111 test samples used in the visual comparison. We crop the cloudy observation, prediction, and paired clean reference from the lossless comparison PNGs and align all RGB images at \(256\times256\) resolution. The images are converted to grayscale using BT.709 coefficients, smoothed using a \(3\times3\) binomial kernel, and processed using Sobel responses followed by four-direction non-maximum suppression. We use an edge threshold of \(0.10\) and a one-pixel matching tolerance. Let edge precision and recall be computed from the matched prediction and reference edges. We report \begin{equation} \mathrm{PredOnly} = 1-\mathrm{EdgePrecision}, \qquad \mathrm{Missing} = 1-\mathrm{EdgeRecall}, \end{equation} together with their harmonic mean, \begin{equation} \mathrm{EdgeF1} = \frac{ 2\,\mathrm{EdgePrecision}\,\mathrm{EdgeRecall} }{ \mathrm{EdgePrecision}+\mathrm{EdgeRecall} }. \end{equation} Dataset-level values are obtained by averaging the per-sample measurements. Confidence intervals use 2,000 paired bootstrap resamples. Because this audit operates on saved aligned 8-bit RGB outputs, it characterizes structural failure profiles and does not replace the official radiometric benchmark evaluation reported in the main paper. \subsection{Source-Wise Scene-Correspondence Intervention} \label{app:source_intervention} For each auxiliary evidence source, the matched condition is compared with a same-modality condition sampled from another scene: \begin{itemize} \item NIR: aligned current-scene NIR versus NIR from another scene; \item spectral: paired non-RGB optical evidence versus spectral evidence from another scene; \item SAR: paired SAR evidence versus SAR evidence from another scene; \item temporal: auxiliary observations from the current scene versus observations from another scene. \end{itemize} The RGB anchor observation, model parameters, token count, and initial generation noise remain fixed. Unlike zero masking, this intervention preserves the modality type and approximate marginal input distribution while breaking correspondence with the current scene. We define \begin{equation} \Delta\mathrm{PSNR} = \mathrm{PSNR}_{\mathrm{matched}} - \mathrm{PSNR}_{\mathrm{shuffled}}, \end{equation} and \begin{equation} \Delta\mathrm{EdgeF1} = \mathrm{EdgeF1}_{\mathrm{matched}} - \mathrm{EdgeF1}_{\mathrm{shuffled}}. \end{equation} For LPIPS and SAM, where lower values indicate better performance, we report the matched-to-shuffled change directly without reversing its direction. Confidence intervals are computed over matched test samples using paired bootstrap. For SEN12MS-CR, the bootstrap unit is the ROI cluster. \subsection{Intervention-Derived Spatial Attribution} \label{app:spatial_attribution_protocol} For source \(s\) and spatial token or patch group \(k\), we replace the corresponding region with the same-position region from a foreign scene, while keeping all other observations and the initial generation noise fixed. Let \(\hat{\mathbf{x}}\) denote the original prediction and \(\hat{\mathbf{x}}^{(s,k)}\) the prediction after intervention. We define the attribution score as the reference-free output discrepancy \begin{equation} a_{s,k} = D\!\left( \hat{\mathbf{x}}, \hat{\mathbf{x}}^{(s,k)} \right). \label{eq:app_intervention_attribution} \end{equation} The paired clean target is not used to construct the ranking. Thus, the subsequent faithfulness evaluation does not rank patches using the same ground-truth metric with which the ranking is evaluated. \subsection{Attribution Faithfulness} \label{app:attribution_faithfulness} Spatial evidence groups are ranked according to Eq.~\eqref{eq:app_intervention_attribution}. We then replace either the top-ranked or randomly selected \(10\%\), \(20\%\), \(30\%\), and \(40\%\) of the evidence groups using the same foreign-scene replacement operator. For a higher-is-better metric \(M\), such as PSNR or Edge F1, the AOPC is \begin{equation} \mathrm{AOPC}_{M} = \frac{1}{4} \sum_{q\in\{0.1,0.2,0.3,0.4\}} \left[ M(\hat{\mathbf{x}}) - M(\hat{\mathbf{x}}_{q}) \right], \end{equation} where \(\hat{\mathbf{x}}_{q}\) is the prediction after replacing fraction \(q\) of the evidence groups. A larger AOPC indicates that the intervened groups exert greater functional influence on reconstruction. Faithfulness is evaluated by comparing the AOPC of top-ranked and random interventions under the same replacement operator. Paired-bootstrap confidence intervals are reported for their difference. The spatial faithfulness study covers NIR, SAR, and temporal evidence; it does not claim patch-level attribution results for the spectral source.

%%%%%%%%%%%%%%%%%%%%%%%%%%%%%%%%%%%%%%%%%%%%%%%%%%%%%%%%%%%%
\section{Full Condition Ablations}
\label{app:full_condition_ablations}

The main paper focuses on how heterogeneous evidence affects structural
fidelity. For completeness, Table~\ref{tab:app_full_ablation} reports the
original reconstruction metrics for the corresponding condition ablations.

\begin{table*}[t]
\centering
\caption{
Full condition ablations of \method{}.
``Spec.'' denotes non-RGB spectral optical bands.
}
\label{tab:app_full_ablation}
\small
\setlength{\tabcolsep}{4pt}

\textbf{(a) CUHK-CR1 and CUHK-CR2}
\vspace{0.2em}

\begin{tabular}{lcccccc}
\toprule
\multirow{2}{*}{Condition}
&
\multicolumn{3}{c}{CUHK-CR1}
&
\multicolumn{3}{c}{CUHK-CR2}
\\
\cmidrule(lr){2-4}
\cmidrule(lr){5-7}
&
PSNR$\uparrow$
& SSIM$\uparrow$
& LPIPS$\downarrow$
&
PSNR$\uparrow$
& SSIM$\uparrow$
& LPIPS$\downarrow$
\\
\midrule
RGB only
& 26.722 & 0.7113 & 0.0251
& 23.997 & 0.6629 & 0.0340 \\
RGB + NIR
& \textbf{28.310} & \textbf{0.8170} & \textbf{0.0189}
& \textbf{25.610} & \textbf{0.7100} & \textbf{0.0272} \\
\bottomrule
\end{tabular}

\vspace{0.7em}

\begin{minipage}[t]{0.50\textwidth}
\centering
\textbf{(b) SEN12MS-CR}
\vspace{0.2em}

\begin{tabular}{lcccc}
\toprule
Condition
& PSNR$\uparrow$
& SSIM$\uparrow$
& MAE$\downarrow$
& SAM$\downarrow$ \\
\midrule
RGB only
& 28.620 & 0.705 & 0.045 & 7.180 \\
RGB + Spec.
& 29.180 & 0.794 & 0.032 & 6.840 \\
RGB + Spec. + SAR
& \textbf{33.120}
& \textbf{0.934}
& \textbf{0.016}
& \textbf{4.900} \\
\bottomrule
\end{tabular}
\end{minipage}
\hfill
\begin{minipage}[t]{0.44\textwidth}
\centering
\textbf{(c) Sen2\_MTC\_New}
\vspace{0.2em}

\begin{tabular}{lccc}
\toprule
Temporal length
& PSNR$\uparrow$
& SSIM$\uparrow$
& LPIPS$\downarrow$ \\
\midrule
$L=1$ & 16.220 & 0.501 & 0.432 \\
$L=2$ & 18.350 & 0.641 & 0.327 \\
$L=3$ & \textbf{21.050} & \textbf{0.724} & \textbf{0.230} \\
\bottomrule
\end{tabular}
\end{minipage}

\end{table*}

The conventional reconstruction metrics follow the same trend as the
structural analysis in the main paper. Aligned NIR improves both CUHK
benchmarks; spectral and SAR evidence progressively improve reconstruction on
SEN12MS-CR; and increasing temporal context improves restoration on
Sen2\_MTC\_New.

%%%%%%%%%%%%%%%%%%%%%%%%%%%%%%%%%%%%%%%%%%%%%%%%%%%%%%%%%%%%
\section{Additional Grounding Controls}
\label{app:additional_grounding_controls}

\subsection{Temporal Reliability Controls}
\label{app:temporal_controls}

We further examine the deterministic temporal reliability mechanism using
matched variants trained for 6K steps with three random seeds.

\begin{table}[t]
\centering
\caption{
Matched temporal-conditioning controls on Sen2\_MTC\_New.
Results report mean and standard deviation over three seeds.
}
\label{tab:temporal_controls}
\small
\setlength{\tabcolsep}{4pt}
\resizebox{0.95\linewidth}{!}{
\begin{tabular}{lccc}
\toprule
Temporal conditioning
& PSNR$\uparrow$
& SSIM$\uparrow$
& LPIPS$\downarrow$ \\
\midrule
Full reliability gate
& \textbf{21.10$\pm$0.12}
& \textbf{0.725$\pm$0.005}
& \textbf{0.229$\pm$0.006} \\
Brightness only
& 21.05$\pm$0.13
& 0.724$\pm$0.005
& 0.230$\pm$0.006 \\
Brightness + whiteness
& 21.01$\pm$0.14
& 0.722$\pm$0.005
& 0.232$\pm$0.006 \\
Brightness + both disagreement terms
& 21.03$\pm$0.13
& 0.723$\pm$0.005
& 0.231$\pm$0.006 \\
Raw temporal tokens
& 20.90$\pm$0.18
& 0.720$\pm$0.007
& 0.235$\pm$0.008 \\
\bottomrule
\end{tabular}
}
\end{table}

The full deterministic gate slightly outperforms direct raw-token
conditioning. However, brightness-only weighting is already within the
observed seed variation of the full gate, while the additional cues provide
no meaningful improvement beyond this variability. We therefore do not claim
that every reliability cue is independently necessary; on this benchmark,
brightness accounts for most of the observed benefit.

%%%%%%%%%%%%%%%%%%%%%%%%%%%%%%%%%%%%%%%%%%%%%%%%%%%%%%%%%%%%
\section{Additional Reference-Free Reliability Analysis} \label{app:dispersion_analysis} The main interventional analyses evaluate evidence dependence and attribution faithfulness using controlled source or patch replacement. We additionally examine whether stochastic predictive instability provides a reference-free signal of residual error. For each input, five predictions are generated using independent latent seeds, and predictive dispersion is used as an instability score. The score is compared with reference-derived pixel reconstruction error and structural inconsistency. \begin{table}[t] \centering \caption{ Diagnostic performance of five-seed predictive dispersion. AURC reduction is reported relative to random ranking. } \label{tab:app_dispersion} \small \setlength{\tabcolsep}{4.5pt} \begin{tabular}{lccc} \toprule Target & Spearman $\rho\uparrow$ & AUROC$\uparrow$ & AURC reduction$\uparrow$ \\ \midrule Pixel reconstruction error & 0.38 & 0.71 & 22\% \\ Structural inconsistency & 0.33 & 0.68 & 15\% \\ \bottomrule \end{tabular} \end{table} Predictive dispersion provides moderate diagnostic information for both targets and is weaker for structural inconsistency than for pixel error. It should therefore be interpreted as an instability indicator rather than calibrated confidence or a definitive detector of hallucinated content.

%%%%%%%%%%%%%%%%%%%%%%%%%%%%%%%%%%%%%%%%%%%%%%%%%%%%%%%%%%%%
\section{Additional Failure Analysis}
\label{app:additional_failure_analysis}

\begin{figure*}[t]
    \centering
    \includegraphics[width=\linewidth]{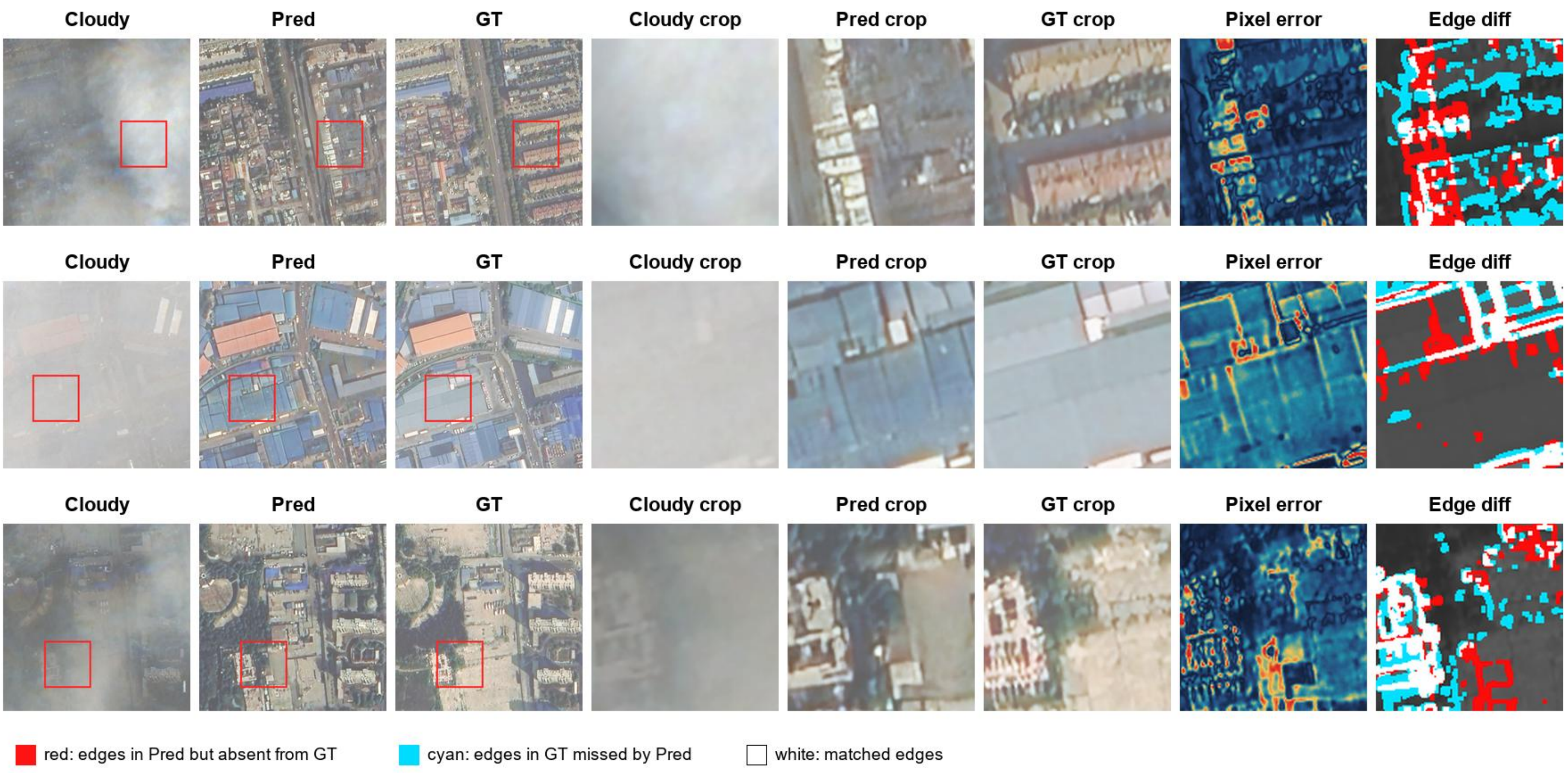}
    \caption{
    Additional failure examples of \method{}.
    Enlarged crops, pixel-error maps, and edge-difference maps reveal local
    radiometric and structural discrepancies between prediction and reference.
    }
    \label{fig:additional_failure_examples}
\end{figure*}

The main paper presents failure analysis across different heterogeneous
observation settings. Figure~\ref{fig:additional_failure_examples} provides
additional examples focusing on local discrepancies between visually coherent
predictions and paired clean references.

The pretrained generative prior can synthesize sharp and coherent land-cover
structures in heavily obscured regions. However, some building-like
boundaries, road-like patterns, or local textures may be absent or spatially
shifted in the paired reference, while some reference structures remain
unrecovered. Pixel-error maps additionally expose local color and radiometric
deviations.

These examples reinforce the distinction between perceptual realism and paired
fidelity. A visually convincing restoration is not necessarily an exact
reconstruction of the underlying scene, and reference-based structural
disagreement should not by itself be interpreted as definitive evidence of
physical hallucination.

%%%%%%%%%%%%%%%%%%%%%%%%%%%%%%%%%%%%%%%%%%%%%%%%%%%%%%%%%%%%
\section{Scope and Limitations} \label{app:limitations} \paragraph{Scope of source-wise attribution.} Scene-correspondence interventions are conducted for NIR, spectral, SAR, and temporal evidence. RGB remains the anchor optical observation and is not subjected to the same matched-versus-shuffled source intervention. Because the auxiliary sources are evaluated in different sensing regimes, their absolute intervention effects should not be interpreted as a universal cross-dataset ranking of modality importance. \paragraph{Scope of spatial attribution.} Patch-level attribution and top-versus-random faithfulness evaluation are conducted for representative NIR, SAR, and temporal sources. We do not claim spatial attribution results for the spectral source. Moreover, the analysis establishes the functional influence of ranked spatial evidence groups, but does not assign semantic interpretations such as a modality attending primarily to a particular land-cover class. \paragraph{Intervention distribution.} Foreign-scene replacement retains the modality type, token count, and approximate marginal distribution and is less distribution-disruptive than zero masking. Nevertheless, replacing an entire source or a local patch may still introduce combinations not observed during training. The results should therefore be interpreted as controlled interventional sensitivity, rather than as proof that all distribution-shift effects have been removed. \paragraph{Failure-profile audit.} The specialist-baseline comparison is a post-hoc structural audit of saved aligned 8-bit RGB outputs. It is used to compare the balance between missing and prediction-only structures and does not replace the official benchmark metrics. Differences in this audit should not be attributed solely to model size, since the compared systems also differ in architecture, objective, training data, and sampling procedure. \paragraph{Reference-based structural diagnostics.} Prediction-only structures, missing structures, and Edge F1 characterize disagreement with paired clean references. Such discrepancies may also reflect registration differences, temporal mismatch, radiometric variation, or ambiguity in the paired target. These diagnostics therefore characterize reference-based structural inconsistency rather than definitive physical hallucination. \paragraph{Residual reliability.} Richer heterogeneous evidence reduces structural inconsistency but does not guarantee physically correct reconstruction. Predictive dispersion provides only moderate diagnostic ability and requires multiple generations per input. Neither the attribution analysis nor the dispersion score constitutes calibrated uncertainty or a replacement for physical validation and expert review.

%%%%%%%%%%%%%%%%%%%%%%%%%%%%%%%%%%%%%%%%%%%%%%%%%%%%%%%%%%%%
\section{Broader Impact}
\label{app:broader_impact}

Cloud removal can increase the availability and usability of optical satellite
imagery for environmental monitoring, agricultural assessment, disaster
response, and land-cover analysis. Incorporating heterogeneous observations
such as optical bands, SAR measurements, and temporal frames can provide
additional scene evidence when cloud contamination limits direct optical
information.

At the same time, generative cloud removal should be used with care in
scientific and decision-making applications. as demonstrated by the failure-profile audit and residual analysis in Sec.~\ref{subsec:evidence_sufficiency}, a pretrained generative
backbone can produce visually plausible but reference-inconsistent structures
in highly ambiguous regions. Restored imagery should therefore not be
interpreted as direct physical measurement or independently verified
observation without appropriate validation or expert review. This is
particularly important for disaster assessment, infrastructure monitoring,
and land-use decision making.

The proposed framework also relies on a large pretrained backbone and GPU
computation. Although \method{} adapts an existing foundation model rather
than training one from scratch, practical deployment should consider the
associated computational and energy costs. Future work should investigate
more efficient adaptation, stronger physical and radiometric constraints, and
better-calibrated reliability estimation.

Finally, remote-sensing technologies are dual-use. Improved cloud removal may
benefit scientific and humanitarian monitoring, but could also enhance
surveillance capabilities when applied to sensitive regions or populations.
We encourage deployment in accordance with applicable data policies, privacy
norms, and domain-specific validation requirements.

\end{document}